\documentclass[letterpaper, 10 pt, conference]{ieeeconf}  
\usepackage{graphicx}   
\usepackage{tabularx} 
\usepackage{array}    
\usepackage{amsmath}    
\usepackage{amssymb}    
\usepackage{amsfonts}   
\usepackage{graphicx} 
\usepackage{tikz}
\usepackage{amsmath}
\usepackage[hidelinks]{hyperref}
\usepackage{url}

\usetikzlibrary{arrows.meta,positioning,fit,calc,shadows.blur,shapes.multipart}
\usepackage{pgfplots}
\pgfplotsset{compat=1.18}
\tikzset{
  >={Latex[length=2mm]},
  box/.style={draw, rounded corners=2mm, thick, align=center, fill=white, blur shadow},
  smallbox/.style={draw, rounded corners=1.2mm, semithick, align=center, fill=white},
  io/.style={draw, dashed, rounded corners=2mm, align=center},
  bubble/.style={draw, rounded corners=1mm, semithick, fill=white, inner sep=2pt},
  edge label/.style={midway, fill=white, inner sep=1pt, font=\footnotesize},
}

\IEEEoverridecommandlockouts                              

\title{\LARGE \bf
Explainable AI-Enhanced Supervisory Control for Robust Multi-Agent Robotic Systems
}

\author{%
Reza Pirayeshshirazinezhad$^{1,3}$ and Nima Fathi$^{2,3}$%
\thanks{This work was supported in part by NASA under Contract NM--NNX15AM73A and Texas A\&M University.}\\[4pt]
\begin{tabular}{c}
$^{1}$Visual Computing and Computational Media, Texas A\&M University, College Station, TX 77843, USA\\
\tt\small \href{mailto:rezapirayesh@tamu.edu}{rezapirayesh@tamu.edu}, \href{mailto:rpirayeshs@gmail.com}{rpirayeshs@gmail.com}
\end{tabular}\\[6pt]
\begin{tabular}{c}
$^{2}$Mechanical Engineering Department, Texas A\&M University, College Station, TX 77843, USA\\
$^{3}$Department of Marine Engineering Technology, Texas A\&M University, Galveston, TX 77553, USA\\
\tt\small \href{mailto:nfathi@tamu.edu}{nfathi@tamu.edu}
\end{tabular}
}

\begin{document}

\maketitle
\thispagestyle{empty}
\pagestyle{empty}

\begin{abstract}
We present an explainable AI–enhanced supervisory control framework for multi-agent robotics that combines (i) a timed-automata supervisor for safe, auditable mode switching, (ii) robust continuous control (Lyapunov-based controller for large-angle maneuver; sliding-mode controller (SMC) with boundary layers for precision and disturbance rejection), and (iii) an explainable predictor that maps mission context to gains and expected performance (energy, error). Monte Carlo–driven optimization provides the training data, enabling transparent real-time trade-offs.

We validated the approach in two contrasting domains, spacecraft formation flying and autonomous underwater vehicles (AUVs). Despite different environments (gravity/actuator bias vs. hydrodynamic drag/currents), both share uncertain six degrees of freedom (DoF)6-DOF rigid-body dynamics, relative motion, and tight tracking needs, making them representative of general robotic systems. In the space mission, the supervisory logic selects parameters that meet mission criteria. In AUV leader–follower tests, the same SMC structure maintains a fixed offset under stochastic currents with bounded steady error. In spacecraft validation, the SMC controller achieved submillimeter alignment with 21.7\% lower tracking error and 81.4\% lower energy consumption compared to Proportional-Derivative PD controller baselines. At the same time, in AUV tests, SMC maintained bounded errors under stochastic currents. These results
highlight both the portability and the interpretability of the approach for
safety-critical, resource-constrained multi-agent robotics.
\end{abstract}

\section{INTRODUCTION}
Precision formation control and cooperative autonomy are critical for multi-agent robotic systems, spanning aerial swarms, underwater vehicles, and space missions. Coordinated formations enable robustness, redundancy, and expanded coverage compared to single platforms, but they also impose strict requirements on guidance, navigation, and control (GNC). Maintaining accurate relative positions and attitudes under uncertain dynamics, disturbances, and limited resources remains a central challenge. In this setting, formation control is inherently multi-agent because the objective is defined over relative states among at least two independently actuated vehicles, and it is realized via distributed (leader–follower or consensus) controllers that use local measurements and limited neighbor/leader information to enforce geometry.

This paper introduces an explainable artificial intelligence (XAI)-enhanced supervisory framework for robust, high-precision formation control. The framework integrates three complementary components: (i) a formally verifiable supervisory layer using timed automata to govern safe and deterministic mission phase transitions, (ii) adaptive robust controllers based on Lyapunov and sliding mode control (SMC) techniques to guarantee stability under noise and disturbances, and (iii) a learning-augmented optimizer that predicts both controller parameters and performance outcomes such as energy consumption and tracking error. Unlike conventional black-box learning approaches, the proposed architecture provides transparency on two levels: locally, the AI predicts not only control gains but also their consequences, and globally, the supervisory logic ensures mission behavior is rule-based and auditable. Safety (in this work) implies invariance of the closed-loop state within mission-certified envelopes (pointing, range, alignment, energy, and time) and deterministic, auditable mode transitions enforced by timed automata, with Lyapunov/SMC controllers ensuring stability and bounded errors under disturbances.

The space mission used in this research is the Virtual Telescope for X-ray Observation (VTXO), a constrained two-spacecraft formation mission. VTXO requires maintaining a ~1 km focal length while maintaining strict accuracy in both translational and rotational dynamics, imposing accuracy, robustness and resource discipline in the guidance, navigation and control (GNC) system. Although our full validation emphasizes this space application, the methodology is designed for broader robotics domains. For generality, we also include an underwater robot formation case study implemented with a robust SMC. Due to the specifics of the AUV formation, no learning or supervisory automation is used; the case serves instead as a control-only baseline, demonstrating that the underlying control laws extend naturally to very different dynamics dominated by hydrodynamic disturbances and sensor noise.

This paper makes the following contributions:
\begin{itemize}
  \item \textbf{Unified XAI Framework:} An explainable adaptive control system that predicts mission parameters, energy consumption, and error, enabling interpretable trade-offs.
  \item \textbf{Formally Verifiable Supervision:} A timed-automata supervisory layer for managing multi-phase missions with deterministic, verifiable transitions.
  \item \textbf{High-Precision Robust Control:} Integration of SMC and Lyapunov-based controller methods for precise relative position and attitude control under noise, with mitigation of chattering.
  \item \textbf{Learning-Based Optimization:} A neural network surrogate for solving constrained, non-convex dynamic optimization problems, enabling rapid gain adaptation.
  \item \textbf{Cross-Domain Demonstration:} Validation of the complete framework in a demanding spacecraft formation scenario, and illustration of baseline control portability in an AUV formation case.
\end{itemize}

Multi-agent coordination has been studied through consensus and
graph-theoretic methods~\cite{OlfatiSaber2007}, while robust
feedback designs such as Lyapunov and sliding-mode control remain the
backbone for safety-critical aerospace and underwater systems~\cite{Khalil2002,
naseri2018formation}. However, these methods are not inherently
explainable or easily integrated with supervisory logic and modern learning
methods, motivating the framework developed in this work.

Beyond spacecraft applications, the framework directly extends to
challenging marine environments. Autonomous underwater vehicles (AUVs),
surface vessels, and cooperative swarms for offshore inspection or disaster
response share the same sources of uncertainty—hydrodynamic drag,
currents, and sensor noise—that were explicitly addressed in our case
study. These maritime domains impose strict safety requirements, for
example in offshore energy, shipping, and search-and-rescue operations,
where robust control and operator trust are critical. By validating the
framework on AUV formations, we illustrate its portability and readiness
for broader marine robotics.
\section{Related Work}
\label{sec:related}
Multi-agent control has been extensively studied across aerial, space, and
underwater domains. Classical formation control methods include consensus-based
laws~\cite{OlfatiSaber2007}, and
leader--follower schemes~\cite{fax2004information}. Robust control strategies
such as SMC ~\cite{Utkin1992,EdwardsSpurgeon1998} and
Lyapunov-based feedback~\cite{Khalil2002} have been applied to spacecraft
attitude~\cite{wie2008space} and underwater vehicle navigation~\cite{fossen2011handbook}.
Recent work has introduced learning-based control for robotics and aerospace,
including supervised learning~\cite{Pan2016}, reinforcement learning~\cite{SuttonBarto2018,Lowe2017},
and deep spacecraft guidance~\cite{Gaudet2020}.  However, such methods are often
black-box in nature, making verification difficult.

Supervisory control has been combined with continuous feedback controllers in
hybrid systems theory~\cite{Cassandras2008}, with timed automata providing a
formalism for reasoning about discrete transitions~\cite{AlurDill1994}.
These structures have been used in robotics~\cite{karaman2011linear} and in
mission sequencing for aerospace applications~\cite{pirayesh2018attitude}.
Our work integrates supervisory logic, robust control, and explainable learning
into a unified framework, demonstrated on spacecraft formation flying and AUV
formations.

\section{System Dynamics}
\label{sec:dynamics}

We consider a generic class of multi-agent robotic systems, each modeled as a
rigid body with six degrees of freedom (6-DOF). The agent states include
position $r \in \mathbf{R}^3$, linear velocity $v \in \mathbf{R}^3$,
attitude represented by a unit quaternion $q \in \mathbf{S}^3$, and angular
velocity $\omega \in \mathbf{R}^3$. For notation, scalars are italic ($x$), vectors are bold lowercase ($\pmb{x}$), and matrices are bold uppercase ($\pmb{X}$). The onboard, estimated, and selected value of a variable $x$ is $\hat{x}$; the measured value is $\tilde{x}$. The operator $\odot$ represents the Hadamard (element-wise) product.
The unified dynamics are

\begin{align}
\dot{\pmb{q}} &= \tfrac{1}{2}\,\pmb{\omega} \otimes \pmb{q}, \label{eq:quat_dyn} \\
\dot{\pmb{\omega}} &= \mathbf{J}^{-1}\!\left(\pmb{\tau} - \pmb{\omega} \times \mathbf{J}\pmb{\omega} + \pmb{d}_{\omega} \right), \label{eq:ang_dyn} \\
\dot{\pmb{r}} &= \pmb{v}, \\
\dot{\pmb{v}} &= \pmb{f}_{\text{env}}(\pmb{r},\pmb{v}) + \pmb{u} + \pmb{d}_v, \label{eq:lin_dyn}
\end{align}

where $\mathbf{J} = \bar{\mathbf{J}} + \Delta \mathbf{J}$ is the inertia matrix with
nominal component $\bar{\mathbf{J}}$ and uncertainty $\Delta \mathbf{J}$,
$\pmb{\tau}$ is the applied control torque, $\pmb{u}$ is the commanded force input,
and $\pmb{d}_{\omega}, \pmb{d}_{v}$ are bounded external disturbances. The function
$f_{\text{env}}(\cdot)$ captures environment-specific effects such as gravitational
forces in space or hydrodynamic forces underwater.

\subsection{Concept of operations for spacecraft formation}
\label{sec:mission_constraints}
VTXO imposes constraints that govern both supervision and control: (i) \emph{angular resolution} of 55 milli-arcsecond (mas) requires sub-mm transverse alignment at a 1\~km focal length during the science phase; (ii) \emph{pointing accuracy} of a few arcminutes per spacecraft (star-tracker FoV bound); (iii) \emph{range keeping} at $1$~km~$\pm$~$1$~m; and (iv) \emph{transient duration} of only a few minutes before entering science mode. The automaton enforces these phases: after a commissioning period, the formation transitions through a transient phase until science phase accuracy is reached, executes science observations on multiple space objects, and finally decommissions. Verification uses both hard constraints (quaternion normalization, positive gains, bounded durations) and probabilistic checks (e.g., $P_{99}(e)$ within FoV). The sensing stack—IMU ($\tilde{\boldsymbol{\omega}},\tilde{\ddot{\boldsymbol{r}}}$), star tracker ($\tilde{\boldsymbol{q}}$), radio ranging ($\tilde r^{z}_{\mathrm{rel}}$), GPS (for orbit calibration), and astrometric sensor ($\tilde{\boldsymbol{r}}^{xy}_{\mathrm{rel}}$)—is fused by an MEKF to estimate $(\boldsymbol{q},\boldsymbol{\omega},\boldsymbol{r}_{\mathrm{rel}},\boldsymbol{v}_{\mathrm{rel}})$ for feedback. These requirements shape timed-automata guards/timers and motivate Lyapunov-based transient control with SMC refinement during science phases.

\section{Control Framework}
\label{sec:control}

The continuous layer of our architecture is composed of robust controllers
that can be applied to spacecraft and underwater vehicles. We employ
Lyapunov-based control for spacecraft large-angle maneuver and sliding-mode control (SMC) laws, which guarantee
stability under disturbances. The supervisory and AI modules
adapt and tune these controllers as needed. 

\subsection{Spacecraft controllers}

For the large-angle maneuver with the shortest-distance rotation in the transient phase, the following Lyapunov controller can ensure asymptotic stability. 
\begin{equation} \label{tau_in}
\hat{\pmb{\tau}}_{in}=-\hat{k_1}\text{sign}(\partial \hat{q_4})\hat{\pmb{\partial q}_{1:3}}-\hat{k_2}(1 - \hat{\pmb{\partial q}_{1:3}}^T\hat{\pmb{\partial q}_{1:3}})\hat{\pmb{\omega}}
\end{equation}

Using the following candidate Lyapunov function
\begin{equation} \label{V}
V(\pmb{x})=\frac{1}{4}\pmb{\omega}^T\pmb{J}\pmb{\omega}+\frac{1}{2}k_1\pmb{\partial q}^T_{1:3}\pmb{\partial q}_{1:3}+\frac{1}{2}k_2(1-\partial q_4)^2
\end{equation}

For robustness to modeling errors and disturbances, SMC is used when robustness is needed toward disturbances for high-precision formation in the science phase.
For lower energy consumption and deterministic rotation for spacecraft formation, the sliding surface is chosen to provide the shortest-distance rotation. Defining the sliding surface as

\begin{equation} \label{sliding vectorACS}
\hat{\pmb{s}}=(\hat{\pmb{\omega}}-{\pmb{\omega}}_f)+\hat{k}_{SMC}\text{sign}(\partial \hat{q_4})\hat{\pmb{\partial q}_{1:3}}
\end{equation}

Using this optimal surface, the torque input is
\begin{align} 
\begin{split}
\hat{\pmb{\tau}}_{in} = & \pmb{J}\Bigg\{\frac{\hat{k}_{SMC}}{2}\bigg[|\partial \hat{q_4}|({\pmb{\omega}}_f-\hat{\pmb{\omega}}) \\
& -\text{sign}(\partial \hat{q_4})\hat{\pmb{\partial q}_{1:3}} \times ({\pmb{\omega}}_f+\hat{\pmb{\omega}})\bigg] \\
& +\dot{\pmb{\omega}}_f- \hat{\pmb{Z}}\text{sat}(\hat{s_i},\hat{\varepsilon}_i)\Bigg\} \\
& +\hat{\pmb{\omega}}\times\pmb{J}\hat{\pmb{\omega}}
\end{split}
\label{slidingModeU}
\end{align}
where $\hat{k}_{SMC},\hat{\pmb{Z}}>0$ are the estimated controller gains and $\epsilon$ sets the boundary layer. Instead of the discontinuous sign function, which induces chattering, a saturation function is applied to each sliding surface component $s_i$. The function smoothly scales $s_i$ within a boundary layer of thickness $\varepsilon_i$, acting linearly when $|s_i| \leq \varepsilon_i$ and saturating to $\pm 1$ outside. This ensures robustness while reducing chattering.

This structure provides finite-time convergence and disturbance rejection.

\subsubsection{Sliding-Mode Control for Relative Position} \label{SMCR}
We consider a leader--follower formation where the leader drifts (science phase), and the follower regulates the relative position with respect to the leader. Let the relative position and velocity be
\[
\pmb{r}_{\mathrm{rel}}=\pmb{r}_{L}-\pmb{r}_{F}, \qquad 
\dot{\pmb{r}}_{\mathrm{rel}}=\pmb{\text{v}}_{L}-\pmb{\text{v}}_{F},
\]
with desired references \(\pmb{r}^{\,f}_{\mathrm{rel}}\) and \(\pmb{\text{v}}^{\,f}_{\mathrm{rel}}\). The sliding variable for translation is
\begin{equation}\label{eq:s_rel}
\hat{\pmb{s}} = \big(\hat{\pmb{\text{v}}}_{\mathrm{rel}} - \pmb{\text{v}}^{\,f}_{\mathrm{rel}}\big) + \hat{k}\,\big(\hat{\pmb{r}}_{\mathrm{rel}} - \pmb{r}^{\,f}_{\mathrm{rel}}\big),
\end{equation}
where \(\hat{k}>0\) is a gain. During follower control, the assumed relative model \(\bar{f}\) is
\begin{equation}\label{eq:fbar_rel}
\bar{f}\!\big(\hat{\pmb{r}}_{\mathrm{rel}},\hat{\pmb{\text{v}}}_{\mathrm{rel}}\big) \;=\; -\,\mu\!\left(\frac{\hat{\pmb{r}}_{F}}{\|\hat{\pmb{r}}_{F}\|^{3}}-\frac{\hat{\pmb{r}}_{L}}{\|\hat{\pmb{r}}_{L}\|^{3}}\right) - \hat{\pmb{u}}_{\mathrm{in}L},
\end{equation}
so that the follower treats the leader’s thrust \(\hat{\pmb{u}}_{\mathrm{in}L}\) as a disturbance. For natural drift of the leader, \(\hat{\pmb{u}}_{\mathrm{in}L}=\pmb{0}\). In station-keeping we take \(\pmb{\text{v}}^{\,f}_{\mathrm{rel}}=\pmb{0}_{3\times 1}\) and \(\dot{\pmb{\text{v}}}^{\,f}_{\mathrm{rel}}=\pmb{0}_{3\times 1}\).

Imposing the sliding dynamics \(\dot{\hat{\pmb{s}}}=\pmb{0}\) with a boundary-layer approximation yields the follower control law
\begin{equation}\label{eq:uF}
\begin{aligned}
\hat{\pmb{u}}_{\mathrm{in}F} =\;& 
\mu\!\left(\frac{\hat{\pmb{r}}_{F}}{\|\hat{\pmb{r}}_{F}\|^{3}}
      - \frac{\hat{\pmb{r}}_{L}}{\|\hat{\pmb{r}}_{L}\|^{3}}\right)
+ \hat{\pmb{u}}_{\mathrm{in}L} \\[4pt]
&- \hat{k}\,\big(\hat{\pmb{\text{v}}}_{\mathrm{rel}} 
      - \pmb{\text{v}}^{\,f}_{\mathrm{rel}}\big)
- \hat{\pmb{Z}}\,\mathrm{sat}_\varepsilon(\hat{\pmb{s}}).
\end{aligned}
\end{equation}

where \(\hat{\pmb{Z}} \succ \pmb{0}\) is diagonal and \(\mathrm{sat}_\varepsilon(\cdot)\) is applied componentwise:
\[
\mathrm{sat}_\varepsilon(\hat{s}_i)=
\begin{cases}
\;1, & \hat{s}_i>\hat{\varepsilon}_i,\\[2pt]
\;\hat{s}_i/\hat{\varepsilon}_i, & |\hat{s}_i|\le \hat{\varepsilon}_i,\\[2pt]
-1, & \hat{s}_i<-\hat{\varepsilon}_i,
\end{cases}
\qquad i\in\{1,2,3\},
\]
with boundary-layer thickness \(\hat{\varepsilon}_i>0\). The saturation replaces the discontinuous signum to mitigate chattering.

\paragraph*{Lyapunov stability.}
With \(V(\Delta(\pmb{x}))=\tfrac12\,\hat{\pmb{s}}^{\!\top}\hat{\pmb{s}}\), we have \(\dot V=\hat{\pmb{s}}^{\!\top}\dot{\hat{\pmb{s}}}\). Substituting \eqref{eq:uF} into the closed-loop relative dynamics gives $\dot V \;\le\; -\,\lambda_{\min}(\hat{\pmb{Z}})\,\big\|\mathrm{sat}_\varepsilon(\hat{\pmb{s}})\big\|_2^2
\;\le\; 0,$
so \(\hat{\pmb{s}}\to \pmb{0}\) and \((\hat{\pmb{r}}_{\mathrm{rel}},\hat{\pmb{\text{v}}}_{\mathrm{rel}})\) converge to a neighborhood set by \(\hat{\varepsilon}_i\). Thus, the closed loop is globally asymptotically stable to the boundary layer and practically asymptotically stable for small \(\hat{\varepsilon}_i\).

\subsection{AUV Formation Dynamics and Control}

For autonomous underwater vehicles (AUVs), uncertainties are from added mass,
damping coefficients, and environmental variations. The 6-DOF rigid–body dynamics are
\begin{equation}\label{HydroMassNu}
\big(\pmb{M}_{RB} + \pmb{M}_{A}\big)\dot{\pmb{\nu}} 
+ \pmb{C}(\pmb{\nu})\,\pmb{\nu} 
+ \pmb{D}(\pmb{\nu})\,\pmb{\nu} 
= \pmb{\tau} ,
\end{equation}
where the velocity vector is $\pmb{\nu} =
\begin{bmatrix}
\pmb{v}^\top & \pmb{\omega}^\top
\end{bmatrix}^\top ,$ with $\pmb{v}\in\mathbb{R}^3$ the linear velocity and $\pmb{\omega}\in\mathbb{R}^3$ the angular velocity in the body frame.  
Here $\pmb{M}_{RB}$ is the rigid–body mass–inertia matrix, $\pmb{M}_{A}$ is the added–mass matrix, $\pmb{C}(\pmb{\nu})$ captures Coriolis and centripetal effects, and $\pmb{\tau}$ is the generalized input of forces and moments generated by propellers and fins.  

The damping term $\pmb{D}(\pmb{\nu})\,\pmb{\nu}$ is modeled as the sum of linear and quadratic drag components:
\begin{align}
\pmb{M}_{\text{drag}}(\pmb{\omega}) &= \pmb{D}_\omega^{(1)}\,\pmb{\omega} 
  + \pmb{D}_\omega^{(2)}\big(|\pmb{\omega}|\odot \pmb{\omega}\big), \\
\pmb{F}_{\text{drag}}(\pmb{v}_r) &= \pmb{D}_v^{(1)}\,\pmb{v}_r 
  + \pmb{D}_v^{(2)}\big(|\pmb{v}_r|\odot \pmb{v}_r\big),
\end{align}
where $\pmb{v}_r = \pmb{v} - \pmb{R}(q)\pmb{v}_c$ is the velocity relative to the ocean current $\pmb{v}_c$, expressed in the body frame through the quaternion-derived rotation matrix $\pmb{R}(q)$.  
The diagonal matrices $\pmb{D}_\omega^{(1)}, \pmb{D}_\omega^{(2)}, \pmb{D}_v^{(1)}, \pmb{D}_v^{(2)}$ are identified from hydrodynamic coefficients. Uncertainty enters through $\pmb{M}_A$, drag coefficients, buoyancy mismatch, and stochastic ocean currents modeled as first–order Gauss–Markov processes.  

\paragraph*{Sliding variables}
Attitude error is computed from the quaternion 
$\pmb{q}_e = \pmb{q}_d \otimes \pmb{q}^\ast$, where 
$\pmb{q}_d = [\,1,\;0,\;0,\;0\,]^\top$. 
Its vector part $\pmb{q}_{e,v}$ and scalar part $q_{e,0}$ yield the 
rotational sliding surface.

\begin{equation}
\pmb{s}_\omega = (\pmb{\omega}-\pmb{\omega}_d) + \Lambda_\omega \,\mathrm{sgn}(q_{e,0})\,\pmb{q}_{e,v}, \quad \Lambda_\omega\succ 0.
\end{equation}
For translation, with desired trajectory $(\pmb{r}_d,\pmb{v}_d)$, the error and sliding variable are
\begin{equation}
\pmb{e}_r = \pmb{r}-\pmb{r}_d,\;\; \pmb{e}_v=\pmb{v}-\pmb{v}_d,\;\; 
\pmb{s}_v=\pmb{e}_v+\Lambda_v \pmb{e}_r .
\end{equation}

\paragraph*{Control laws.}
The rotational control cancels gyroscopic and drag terms, then adds SMC feedback:
\begin{equation}
\pmb{\tau} = \pmb{\omega}\times \pmb{J}\pmb{\omega} 
+ \pmb{M}_{\text{drag}}(\pmb{\omega}) 
- K_\omega \pmb{s}_\omega 
- B_\omega\,\mathrm{sat}_\varepsilon(\pmb{s}_\omega),
\end{equation}
with $K_\omega,B_\omega\succ0$.  

The translational control adds feed–forward drag, Coriolis, buoyancy, and weight compensation, with SMC feedback:
\begin{equation}
\begin{aligned}
\pmb{u} =\;& \pmb{F}_{\text{drag}}(\pmb{v}_r) 
+ \pmb{M}_{\text{tot}}(\pmb{\omega}\times \pmb{v}_r) 
+ \pmb{F}_b - m g\,\pmb{e}_3 \\[4pt]
&- K_v \pmb{s}_v 
- B_v\,\mathrm{sat}_\varepsilon(\pmb{s}_v).
\end{aligned}
\end{equation}

Here, $\pmb{e}_3 = [0 \;\; 0 \;\; 1]^\top$ is the unit vector along the vertical axis (used to represent gravity), and $\pmb{F}_b = \rho V g \,\pmb{e}_3$ is the buoyancy force vector acting upward, where $\rho$ is the fluid density, $V$ is the displaced volume, and $g$ is the gravitational acceleration.
 $\mathrm{sat}_\varepsilon(\cdot)$ is the componentwise saturation function
\[
\mathrm{sat}_\varepsilon(x_i)=
\begin{cases}
1, & x_i>\varepsilon_i,\\
x_i/\varepsilon_i,& |x_i|\le \varepsilon_i,\\
-1,& x_i<-\varepsilon_i,
\end{cases}
\]
which reduces chattering by linearly scaling errors within a boundary layer $\varepsilon$.  

\paragraph*{Leader–follower structure.}
The leader vehicle tracks an absolute reference $(\pmb{r}_d,\pmb{v}_d)$.  
The follower tracks a relative offset $\pmb{d}$:
\[
\pmb{r}_d = \pmb{r}_L + \pmb{d},\quad 
\pmb{v}_d = \pmb{v}_L,
\]
using the same control laws. This ensures formation maintenance under ocean current disturbances and model uncertainty.  

\paragraph*{Stability.}
A Lyapunov function 
\[
V=\tfrac12 \pmb{s}_\omega^\top \pmb{J} \pmb{s}_\omega + \tfrac12 \pmb{s}_v^\top \pmb{M}_{\text{tot}} \pmb{s}_v
\]
yields
\[
\dot V \le -\lambda_{\min}(K_\omega)\|\pmb{s}_\omega\|^2 
           - \lambda_{\min}(K_v)\|\pmb{s}_v\|^2 
           + \mathcal{O}(\varepsilon),
\]
so errors converge to a neighborhood defined by the boundary layer thickness. This guarantees robust practical stability despite hydrodynamic uncertainty.

\section{Unified Explainable AI Framework (General, Instantiated on VTXO)}
\label{sec:ai_framework}

We present a unified framework for multi–agent robotics in which optimization,
Monte Carlo (MC) simulation, and supervised learning cooperate to adapt control
across mission phases. The learning block is represented as a parametric
function $f_{\theta}$, where $\theta$ denotes trainable weights obtained offline
by stochastic optimizers. The framework is domain–agnostic: dynamics, control
maps, and constraints enter through a \textit{domain adapter}
$\mathcal{D}=(f_{\text{dom}},G_{\text{dom}},\mathcal{C}_{\text{dom}})$. In this
paper we instantiate $\mathcal{D}$ for the VTXO spacecraft formation.

\vspace{2pt}
\subsection{Phase Coupling via Data Generation}
For each phase (e.g., transient, science), MC samples initial states and
uncertainties, then solves a phase–wise tuning problem. Optimization
(Simulated Annealing, Genetic Algorithms) minimizes energy $E$ and error $e$
by adjusting controller gains. The final state of a phase becomes the next
phase's initial state:
\[
\pmb{x}^{(i+1)}_0 \gets \pmb{x}^{(i)}_f.
\]
This coupling yields consistent, phase-linked training data.

\vspace{2pt}
\subsection{Unified Optimization (Domain–Agnostic)}
Define the cost vector
\begin{equation}
\chi = [E,\, e], 
\qquad 
E = \int \pmb{\tau}^\top \pmb{\omega}\, dt, 
\quad 
e = \| \pmb{q} - \pmb{q}_f \|.
\end{equation}
The tuning problem is
\begin{equation}
\min_{\pmb{k},\, T} \;\chi(\pmb{x}_0,\, w,\, \pmb{x}_f,\, \pmb{k})
\end{equation}
subject to domain adapter constraints:
\begin{align}
\dot{\pmb{x}} &= f_{\text{dom}}(\pmb{x}(t),\pmb{\tau}), \\
\hat{\pmb{\tau}} &= G_{\text{dom}}(\pmb{k},\pmb{x}(t)), \\
(\pmb{x}_0,\pmb{x}_f,\pmb{k},T,w) &\in \mathcal{C}_{\text{dom}} .
\end{align}

\paragraph*{VTXO Instantiation.}
For spacecraft, $f_{\text{dom}}$ is rigid–body quaternion dynamics,
$G_{\text{dom}}$ is Lyapunov/SMC torque, and $\mathcal{C}_{\text{dom}}$ enforces
mission constraints: quaternion normalization, gain positivity, phase duration
bounds, state accuracy limits. Initial/final states and weights are sampled
from uniform and Gaussian distributions as listed in Table~\ref{tab:params}. All the parameters, including the sensor parameters and the orbit parameters for the space mission, are given in \cite{pirayeshshirazinezhad2022artificial, pirayeshshirazinezhad2022designing}.

\vspace{2pt}
\subsection{Learning and Deployment}
The supervised learning block is defined as
\begin{equation}
\pmb{y} = f_{\theta}(\pmb{x}_0, w, \pmb{x}_f),
\qquad 
\pmb{y} = [E,\, e,\, k_1,\, k_2,\, T].
\end{equation}
The parameters $\theta$ are optimized offline to minimize prediction error
between $f_{\theta}$ and optimization/MC outputs, using gradient–based
optimizers (e.g., Adam, Nadam). At runtime, $f_{\theta}$ predicts both
controller parameters, duration of transient phase $T$, and expected performance, enabling real–time trade–off
selection under the supervisor.

\vspace{2pt}
\subsection{Explainability and Generality}
\begin{itemize}
\item \textbf{Predictive transparency:} $f_{\theta}$ outputs $(E,e)$ along with the
controller parameters, making the energy-accuracy trade-offs explicit.  
\item \textbf{Structural transparency:} Timed–automata supervisory logic
governs phase sequencing, ensuring determinism and verifiability.  
\item \textbf{Generality:} The pipeline depends only on $\mathcal{D}$. Changing
$(f_{\text{dom}},G_{\text{dom}},\mathcal{C}_{\text{dom}})$ adapts the same
procedure to UAVs, AUVs, or ground robots without altering $f_{\theta}$.
\end{itemize}

\begin{table}[t]
\centering
\caption{VTXO Parameters.}
\label{tab:params}
\begingroup
\setlength{\tabcolsep}{4pt}           
\renewcommand{\arraystretch}{1.25}    
\scriptsize                           
\begin{tabularx}{\columnwidth}{@{} l >{\raggedright\arraybackslash}X @{}}
\hline
\textbf{Parameter} & \textbf{Value / Range (VTXO)} \\
\hline
Inertia matrix $J$                 & $\mathrm{diag}(1.2,\,1.3,\,0.9)\ \mathrm{kg\cdot m^2}$ \\
Control gains $k_1,k_2$           & $0.01 < k < 3$ \\
Phase duration $T$                 & $7.2$–$72\ \mathrm{s}$ \\
Quaternion norm                    & $\pmb{q}_0^\top \pmb{q}_0 = 1$ \\
Initial angular velocity $\omega_0$ & $N(\mu_1,\sigma_1),\ j\!=\!5,6,7$ \\
Final angular velocity $\omega_f$   & $N(\mu_2,\sigma_2),\ j\!=\!5,6,7$ \\
Initial quaternion components $q_{0j}$ & $U(r^l_j,r^u_j),\ j\!=\!1,2,3,4$ \\
Final quaternion components $q_{fj}$   & $U(r^l_j,r^u_j),\ j\!=\!1,2,3,4$ \\
Weight distribution $w$            & $\big[N(\mu_3,\sigma_3),\ U(c_1^l,c_1^u),\ U(c_2^l,c_2^u)\big]$ \\
Disturbance model                  & Bounded torque bias, sensor noise \\
\hline
\end{tabularx}
\endgroup
\end{table}

\begin{figure}[t]
\centering
\begin{tikzpicture}[
  node distance=10mm, font=\small,
  box/.style={
    rectangle, draw, rounded corners,
    minimum width=22mm, minimum height=8mm,
    text width=28mm, align=center, inner sep=2pt, fill=gray!10
  }
]
\node[box] (opt) {Optimization\\ \scriptsize with Monte Carlo};
\node[box, right=25mm of opt] (ml) {$f_{\theta}$\\ \scriptsize (learned)};
\node[box, below=12mm of ml] (sup) {Supervisor\\ \scriptsize (Timed Automata)};
\node[box, left=25mm of sup] (dom) {Domain Adapter\\ \scriptsize $(f_{\text{dom}},\,G_{\text{dom}},\,\mathcal{C}_{\text{dom}})$};

\draw[->, thick, shorten >=2pt, shorten <=2pt]
  (opt.east) -- (ml.west)
  node[midway, above, font=\scriptsize]{training data};

\draw[->, thick, shorten >=2pt, shorten <=2pt] 
  (ml.south) -- (sup.north)
  node[pos=0.25, anchor=east, font=\scriptsize]{parameters + performance};

\draw[->, thick, shorten >=2pt, shorten <=2pt]
  (sup.west) -- (dom.east);

\draw[->, thick, shorten >=2pt, shorten <=2pt]
  (dom.north) -- (opt.south);
\end{tikzpicture}

\caption{Unified explainable AI framework. Optimization with Monte Carlo (MC)
generates data; $f_{\theta}$ predicts controller parameters and performance; the
supervisor enforces phase sequencing; the domain adapter encodes dynamics,
control laws, and constraints. }
\label{fig:framework}
\end{figure}
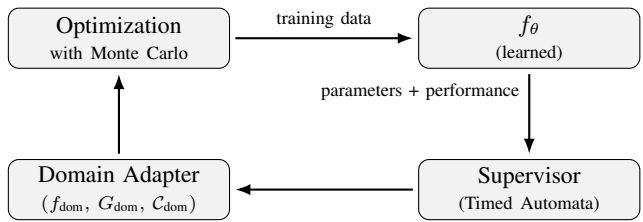

While the framework leverages learning-based optimization, explainability
is central to its design. At runtime, the surrogate $f_{\theta}$ does not merely
output controller gains but also their expected consequences in terms of
energy expenditure and terminal alignment accuracy. These predictions
enable operators and mission planners to understand the trade-offs before
deployment, supporting informed decisions and mission assurance.

Moreover, transparency operates at two levels: (i) locally, the gain–error–
energy mappings are interpretable through performance envelopes that can
be visualized and audited, and (ii) globally, the timed-automata supervisor
provides rule-based and formally verifiable phase sequencing. Together,
these mechanisms reduce operator workload, facilitate audit trails for
safety certification, and enable accountability in safety-critical missions
such as satellite formation flying or underwater search-and-rescue.

\section{Supervisory Adaptive Control System}

The supervisory layer is modeled using timed automata
\cite{hendriks2006model,paxman2004switching}, which captures
discrete mission phases and the conditions for switching between
them. Stability of hybrid supervisory/continuous controllers is
well established in the literature \cite{paxman2004switching}.


\begin{table}[t]
\caption{Finite automaton operational modes ($u_3$ rule).}
\label{Objects1}
\setlength{\tabcolsep}{3pt}
\centering
\begin{tabular}{|c|l|c|}
\hline
Id & Mode & Duration \\
\hline
0 & Commissioning & 60 days \\
1 & Sco X-1 & 0.2 hr \\
2 & GX 5-1 & 1.5 hr \\
3 & GRS 1915+105 & 4.2 hr \\
4 & Cyg X-3 & 4.9 hr \\
5 & Crab Pulsar & 5.4 hr \\
6 & Cen X-3 & 19 hr \\
7 & $\gamma$ Cas & 146 hr \\
8 & Eta Carinae & 452 hr \\
9 & Decommissioning & 5 days \\
\hline
\end{tabular}
\end{table}

\subsection{Timed Automata Formalism}
The supervisory automaton is 
$\pmb{TA}=\{\pmb{s},\pmb{t},\pmb{i},\pmb{k},\pmb{Init}\}$, where:

\begin{itemize}
\item $\pmb{s}=\{s_0,\dots,s_6\}$ are discrete states
(commissioning, stabilization, transient, science, next object,
decommissioning, end).
\item $\pmb{t}=\{t_0,\dots,t_5\}$ are local timers tracking
phase durations.
\item $\pmb{i}=\{u_1,u_2,u_3,u_4\}$ are inputs:
$u_1$ stabilization time rule, $u_2$ transient duration
(learned $T$), $u_3$ mission-specific durations from
Table~\ref{Objects1}, $u_4$ time to reach science phase.
\item $\pmb{k}=\{k_0,\dots,k_5\}$ are transition rules, each
defined as
\begin{equation}
k_{ij}=\{s_i,\, g(i,j),\, reset(i,j),\, s_j\},
\end{equation}
with guard conditions $g(i,j)$ (Boolean rules) and reset
functions $reset(i,j)$.
\item $\pmb{Init}$ are initial conditions with all timers set
to zero and $s_0$ active.
\end{itemize}

Here, $i$ and $j$ denote the indices of the current and next states, respectively, so that each transition $k_{ij}$ encodes a move from $s_i$ to $s_j$ under its guard and reset rules.

Tables~\ref{ResetC} and \ref{GC} list representative reset
functions and guard conditions.

\begin{table}[t]
\caption{Reset functions.}
\label{ResetC}
\centering
\begin{tabular}{|c|c|}
\hline
$reset(i,j)$ & Operation \\
\hline
$reset(0,4)$ & $t_0=0$ \\
$reset(1,2)$ & $t_1=0$ \\
$reset(2,3)$ & $t_2=0$ \\
$reset(3,4)$ & $t_1=0$ \\
$reset(3,5)$ & $t_3=0$ \\
$reset(4,2)$ & $t_1=0$ \\
$reset(5,6)$ & $t_5=0$ \\
\hline
\end{tabular}
\end{table}

\begin{table}[t]
\caption{Boolean guard conditions.}
\label{GC}
\centering
\begin{tabular}{|c|c|}
\hline
Guard $g(i,j)$ & Condition \\
\hline
$g(0,4)$ & $u_3 < t_0$ \\
$g(1,2)$ & $u_1 < t_1$ \\
$g(2,3)$ & $u_2 < t_2$ \\
$g(3,4)$ & $u_3 \geq t_3$ \\
$g(3,1)$ & $u_3 < t_3$ \\
$g(3,5)$ & $452\,\text{hr} \leq t_3$ \\
$g(4,2)$ & $u_4 < t_4$ \\
$g(5,6)$ & $u_3 < t_5$ \\
\hline
\end{tabular}
\end{table}

\subsection{Integration with ACS}
The timed automata supervisor orchestrates switching among SMC and Lyapunov-based controllers, ensuring feasibility and mission safety. The learning
module $f_{\theta}$ provides updated transient duration $T$ and
controller gains, enabling adaptivity. SMC guarantees global
asymptotic stability with positive gains for any trajectory, and
the supervisor ensures correct sequencing of mission phases. This
integration extends mission lifetime and satisfies accuracy and
energy requirements.

\begin{figure}[t]
\centering
\includegraphics[width=0.8\columnwidth]{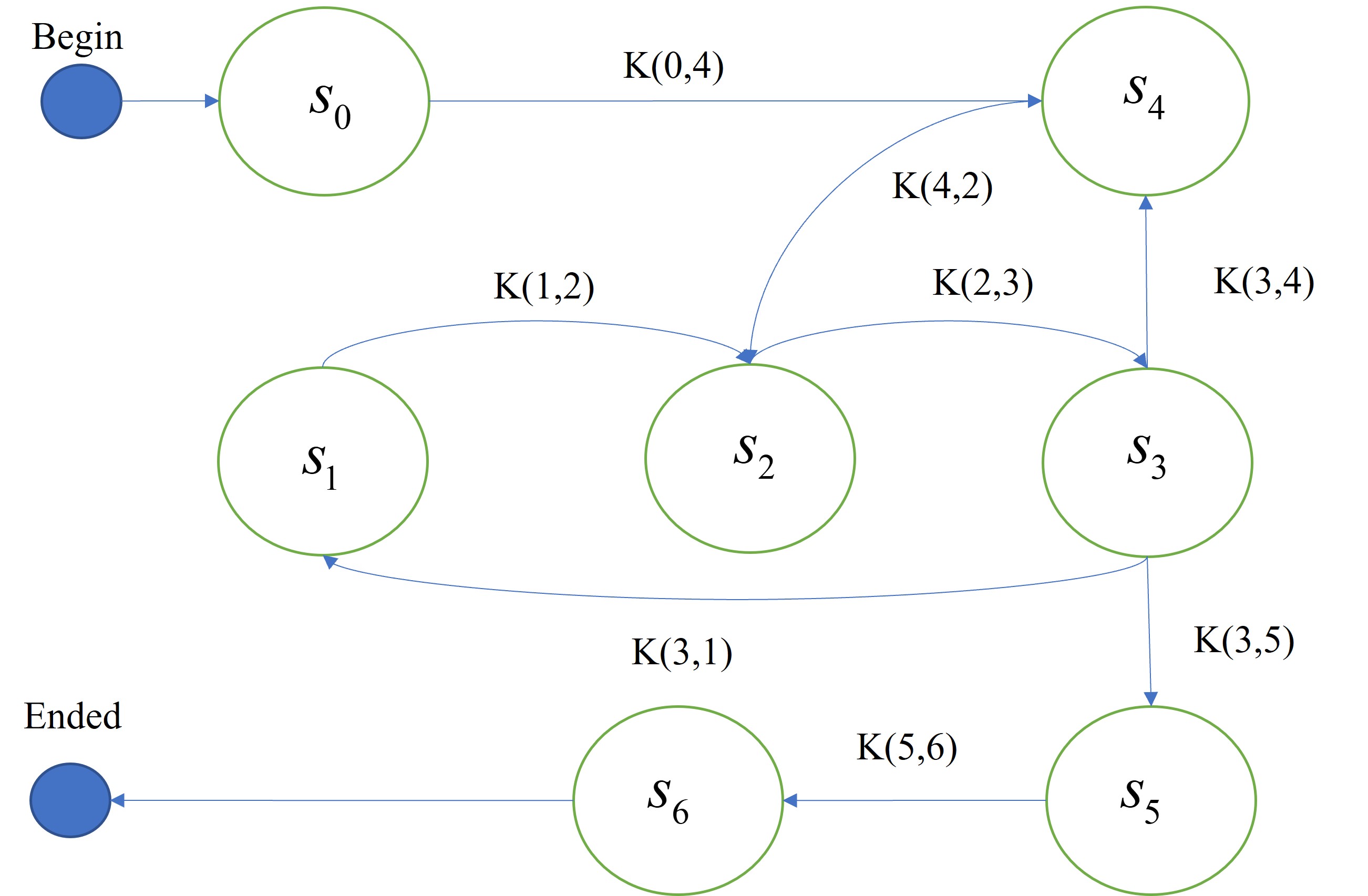}
\caption{Timed automata supervisory control system. States
$\pmb{s}$ represent operational phases; transitions $\pmb{k}$
govern phase switching.}
\label{ST}
\end{figure}

\section{Experiments}
\label{sec:exp}

\textbf{Monte Carlo (MC) pipeline.}
We use MC to (i) assess closed-loop performance/stability, (ii) generate
training data for $f_{\theta}$, and (iii) select controller gains and the
transient duration $T$.


\section{Data Analysis and \texorpdfstring{$f_{\theta}$}{f-theta} for Transient Phase}
\label{sec:transient_data}



\begin{table}[t]
\centering
\caption{Data statistics and DNN predictions (99th percentile).}
\label{tab:stats_dnn}
\setlength{\tabcolsep}{3pt}
\begin{tabular}{|l|c|c|c|c|c|c|}
\hline
$\pmb{y}$ & Mean & Mode & Var & $P_{1}$ & Max & $P_{99}$ / DNN \\
\hline
$k_1$      & 0.1342 & 0.085  & 0.0070 & 0.0244 & 0.4998 & $0.41$/\textbf{0.5627} \\
$k_2$      & 0.2906 & 0.25   & 0.0077 & 0.0926 & 0.4998 & $0.49$/\textbf{0.7580} \\
$e$ (deg)  & 0.0233 & 0.015  & $6.24{\times}10^{-4}$ & 0.0035 & 0.0990 & $0.067$/\textbf{0.0867} \\
$E$ (J)    & 0.1349 & 0.075  & 0.0058 & 0.0229 & 0.6302 & $0.37$/\textbf{0.4695} \\
$T$ (s)    & 48.0   & 72     & 326.46 & 9.0    & 72.0   & $70.53$/\textbf{129.26} \\
\hline
\end{tabular}
\end{table}

Table~\ref{tab:stats_dnn} indicates that maxima of $e$ and $E$ are well above
$P_{99}$ due to solver sensitivity and stochastic SA; the $P_{99}$
bounds confirm robustness for supervisory decisions. As $e\!\to\!0$, the terminal
angular velocity $\pmb{\omega}_f$ concentrates near zero (table \ref{tab:omega_f});
all components satisfy $|P_{99}|\!<\!1\,\mathrm{deg/s}$.

\begin{table}[t]
\centering
\caption{Terminal rate statistics ($\pmb{\omega}_f$).}
\label{tab:omega_f}
\setlength{\tabcolsep}{6pt}
\begin{tabular}{|l|c|c|c|c|}
\hline
 & Mean (deg/s) & Var & $P_{99}$ & $P_{1}$ \\
\hline
$\omega_{f1}$ & $1.03{\times}10^{-4}$ & 0.0027 & 0.1164 & $-0.0972$ \\
$\omega_{f2}$ & $-5.88{\times}10^{-4}$ & 0.0023 & 0.0592 & $-0.1927$ \\
$\omega_{f3}$ & $9.08{\times}10^{-5}$ & 0.0034 & 0.1211 & $-0.0938$ \\
\hline
\end{tabular}
\end{table}


\textbf{Mission compliance.}
Percentile bounds satisfy requirements: $P_{99}(e)=0.067^{\circ}<0.09^{\circ}$
and $P_{99}(E)=0.37\,$J per orbit. SA plays an MPC-like role offline: it finds
feasible, low-energy solutions under hard constraints, at the cost of compute
($\sim$30\,min per optimization).

\subsection{Transient-Phase Stability (ACS)}
\label{sec:transient_stability}

We enforce positivity of gains by using a nonnegative output parameterization
in $f_{\theta}$ (e.g., ReLU at the output), which implies $k_1,k_2>0$. With positive Lyapunov gains, the
deterministic closed-loop is asymptotically stable; under MC disturbances, we
observe bounded errors: $P_{99}(\hat e)<0.09^{\circ}$ and
$P_{99}(\hat E)<0.5$\,J on held-out trials. Together with the percentile
guarantees in Table~\ref{tab:stats_dnn}, this establishes practical stability
for supervisory decisions in the transient phase.

\section{Data Analysis and \texorpdfstring{$f_{\theta}$}{f-theta} for Science Phase}
\label{sec:science_data}

\begin{figure}[t]
\centering
\includegraphics[width=0.85\columnwidth]{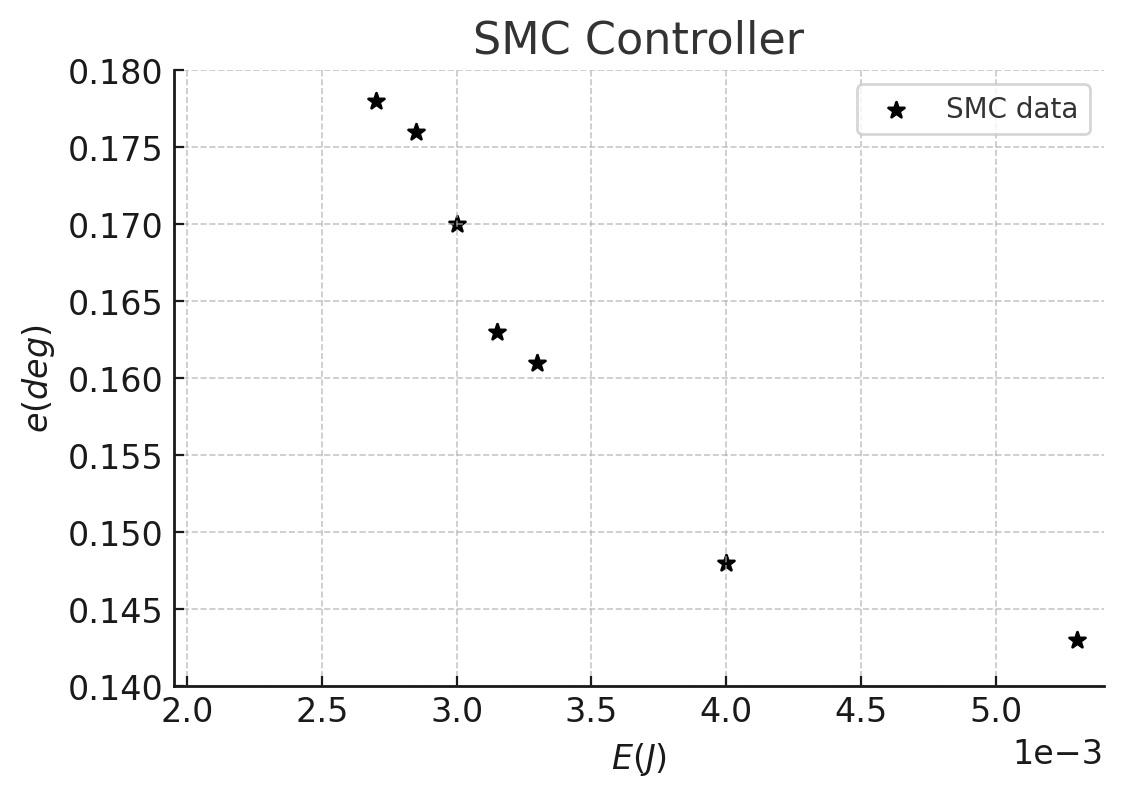}
\caption{Pareto front (SMC, science phase): energy $E$ vs. error $e$.}
\label{SMCGAP}
\end{figure}

For science operations we generate data with a multi-objective GA (MOGA) for
both SMC and PD controllers. A representative SMC Pareto front is shown in
Fig.~\ref{SMCGAP}; similar behavior is observed for PD controller (omitted for space).

\textbf{Learning results.}
A multi-output deep neural network (DNN) for $f_{\theta}$ predicts controller parameters; mean-squared
error (MSE) on held-out sets is $0.0124$ (SMC) vs.\ $0.0434$ (PD), indicating
tighter SMC predictability and superior operating for VTXO.

\subsection{Closed-Loop Simulations (Science)}
\label{sec:science_sims}

We test two weight settings and evaluate the $f_{\theta}$-predicted gains
$\hat{\pmb{k}}$ given in table \ref{tab:k_perf}



\begin{table}[t]
\centering
\caption{Predicted gains $\hat{\pmb{k}}$ and closed-loop performance (science).}
\renewcommand{\arraystretch}{1.25}    
\label{tab:k_perf}
\setlength{\tabcolsep}{6pt}
\begin{tabular}{|l|c|c|l|}
\hline
Controller & $e$ (deg) & $E$ (J) & $\hat{\pmb{k}}$ \\
\hline
PD  & 0.2219 & 0.0353 & $[\,0.1208\;\;0.3786\,]^\top$ \\
SMC & \textbf{0.1738} & \textbf{0.0066} & $[\,2.9947\;\;0.0193\;\;0.2601\,]^\top$ \\
\hline
\end{tabular}
\end{table}

SMC meets the arcminute-level accuracy per spacecraft and yields $>80\%$ lower
energy than PD controller in this setting, while PD controller exceeds the $0.18^{\circ}$ threshold.
This supports SMC as the science-phase controller within the supervisory
framework.

\section{Relative Position Formation (Science)}
\label{sec:relpos}

We consider leader–follower relative motion with disturbances
and sensing via radio ranging ($\tilde r^z_{\mathrm{rel}}$), IMU
($\tilde{\pmb v}_{\mathrm{rel}}$), and interferometry
($\tilde{\pmb r}^{xy}_{\mathrm{rel}}$). PD controller uses $(P,D)=(1,1)$. SMC uses
$\mathrm{sat}(\cdot)\!=\!\mathrm{sign}(\cdot)$ and $k=1$, $\pmb Z=\pmb 1$ per
axis. The references and initial conditions are
\begin{align}
\pmb v^{\,f}_{\mathrm{rel}} &= \pmb 0_{3\times1}, \qquad
\pmb r^{\,f}_{\mathrm{rel}} = [\,0\;0\;1\,]^{\top}\;\mathrm{km}, \\
\pmb v^{\,0}_{\mathrm{rel}} &= [\,1\;1\;1\,]^{\top}\;\mathrm{m/s}, \qquad
\pmb r^{\,0}_{\mathrm{rel}} = [\,1\;2\;10\,]^{\top}\;\mathrm{km}.
\end{align}

\begin{figure}[t]
\centering
\includegraphics[width=0.95\columnwidth]{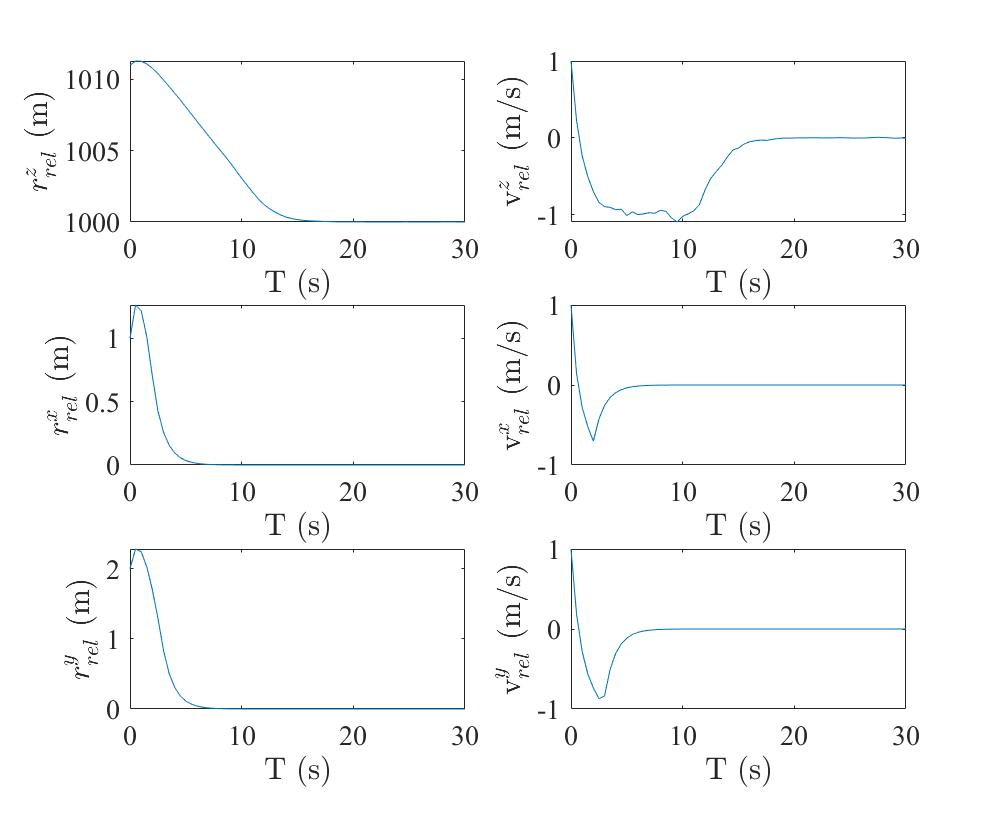}
\caption{Relative position formation with SMC.}
\label{SMCRel}
\end{figure}


SMC achieves the required sub-mm alignment ($\sim 10^{-6}$ scale) despite
disturbances; PD settles around $10^{-3}$ and fails the science tolerance;
Though PD controller augmented with tuned parameters could reduce steady-state error.

\subsection{Controller selection}
During the \emph{transient} phase we employ a Lyapunov-based controller to guarantee global asymptotic stability with smooth, energy-efficient large-angle convergence and low actuator stress. Using SMC here would still converge but typically induces chattering (even with boundary layers), higher effective control effort, and less smooth slews for large initial errors—undesirable at the start of a long mission. In contrast, the science phase demands tight tracking under disturbances for long durations; SMC’s inherent robustness on a properly shaped sliding surface delivers the required precision while tolerating model/torque biases and sensor noise. Hence Lyapunov→SMC is fast, efficient alignment first; then disturbance-rejection and sub-mm accuracy.

\subsection{AUV Formation Experiment}
\label{sec:auv_exp}

\textbf{Setup.}
We simulate a leader--follower AUV pair using the 6-DOF dynamics in \eqref{HydroMassNu} with quaternion attitude and ocean currents modeled as a first–order Gauss–Markov process. The leader tracks a fixed waypoint, and the follower tracks a fixed offset from the leader using the SMC laws. The simulation uses fixed-step RK4. Loosely coupled EKF (as the attitude states are used in the position estimation) is used for fusing the sensors, which are Inertial Navigation System (INS), depth pressure sensors, Sound Navigation and Ranging (Sonar), Doppler Velocity Log (DVL). 

\textbf{Scenario.}
Leader goal: $\pmb{r}_{\text{desired}}^{\,L} = [\,1,\;2,\;3\,]^\top\ \mathrm{m}$, 
$\pmb{v}_{\text{desired}}^{\,L} = \pmb{0}$.
The follower is required to maintain a fixed offset relative to the leader,
$\pmb{d} = [\,3,\;-1,\;1\,]^\top\ \mathrm{m}$, so that
$\pmb{r}_{\text{desired}}^{\,F} = \pmb{r}_{\text{desired}}^{\,L} + \pmb{d}$ and 
$\pmb{v}_{\text{desired}}^{\,F} = \pmb{v}_{\text{desired}}^{\,L}$.
Both vehicles use the same rotational and translational SMC structure with componentwise boundary-layer saturation $\mathrm{sat}_\varepsilon(\cdot)$ to reduce chattering. The values for the AUV simulation are given in Table \ref{tab:auv_all}. The follower uses the relative translational and rotational states in the SMC to follow the leader.
\begin{table}[t]
\centering
\caption{AUV simulation parameters and controller variables.}
\label{tab:auv_all}
\setlength{\tabcolsep}{5pt}
\begin{tabular}{|l|l|}
\hline
\textbf{Symbol} & \textbf{Value} \\ \hline
$\rho$ & $1025$ \\
$V$ & $0.0025$ \\
$m$ & $2.5625$ \\
$m_{a,xx},m_{a,yy},m_{a,zz}$ & $0.36,\;1.00,\;1.50$ \\
$X_u,\;Y_v,\;Z_w$ & $0.048,\;0,\;0.044$ \\
$X_{uu},\;Y_{vv},\;Z_{ww}$ & $5.85,\;11.98,\;21.85$ \\
$K_p,\;M_q,\;N_r$ & $0,\;0,\;21.85$ \\
$\pmb{J}_{\text{add}}$ & $[\,0.001764,\;0.023,\;0.002415\,]$ \\
$\pmb{J}$ & $[\,0.006664,\;0.023,\;0.004515\,]$ \\ \hline
$\pmb{r}_d^{\,L}$ & $[\,1\;\;2\;\;3\,]^\top$ m \\
$\pmb{v}_d^{\,L}$ & $\pmb{0}_{3\times1}$ \\
$\pmb{d}$ & $[\,3\;\;-1\;\;1\,]^\top$ m \\
$\pmb{r}_d^{\,F},\;\pmb{v}_d^{\,F}$ & $\pmb{r}_L+\pmb{d},\;\pmb{v}_L$ \\ \hline
$\Lambda_\omega$ & $\operatorname{diag}(2,2,2)$ \\
$\pmb{K}_\omega,\;\pmb{B}_\omega$ & $\operatorname{diag}(0.8),\;\operatorname{diag}(0.2)$ \\
$\pmb{\varepsilon}_\omega$ & $[\,0.05\;\;0.05\;\;0.05\,]^\top$ rad/s \\ \hline
$\Lambda_v$ & $\operatorname{diag}(1,1,1)$ \\
$\pmb{K}_v,\;\pmb{B}_v$ & $\operatorname{diag}(4),\;\operatorname{diag}(0.8)$ \\
$\pmb{\varepsilon}_v$ & $[\,0.05\;\;0.05\;\;0.05\,]^\top$ m/s \\ \hline
\end{tabular}
\end{table}

\begin{figure}[t]
  \centering
  \includegraphics[width=\columnwidth]{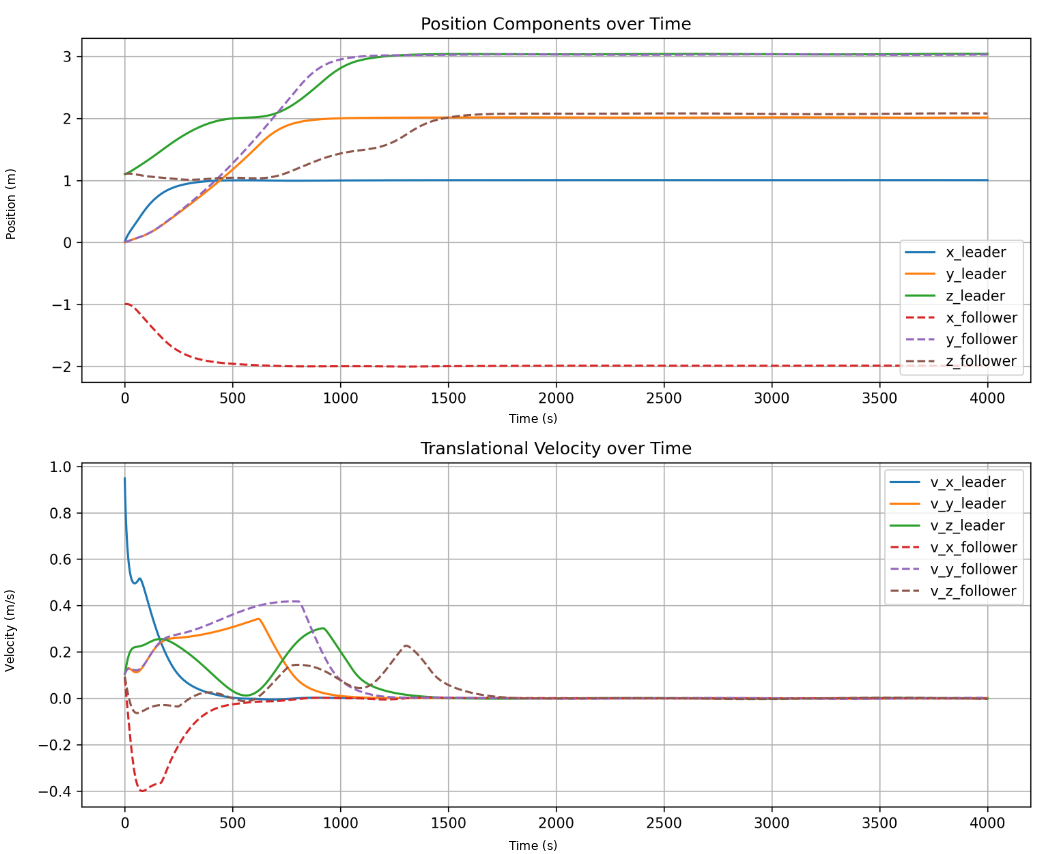}
  \caption{%
  \textbf{AUV translational dynamics.} 
  The follower tracks a fixed relative offset to the leader while rejecting 
  disturbances.
  }
  \label{fig:trans_dynamics}
\end{figure}

\begin{figure}[t]
  \centering
  \includegraphics[width=\columnwidth]{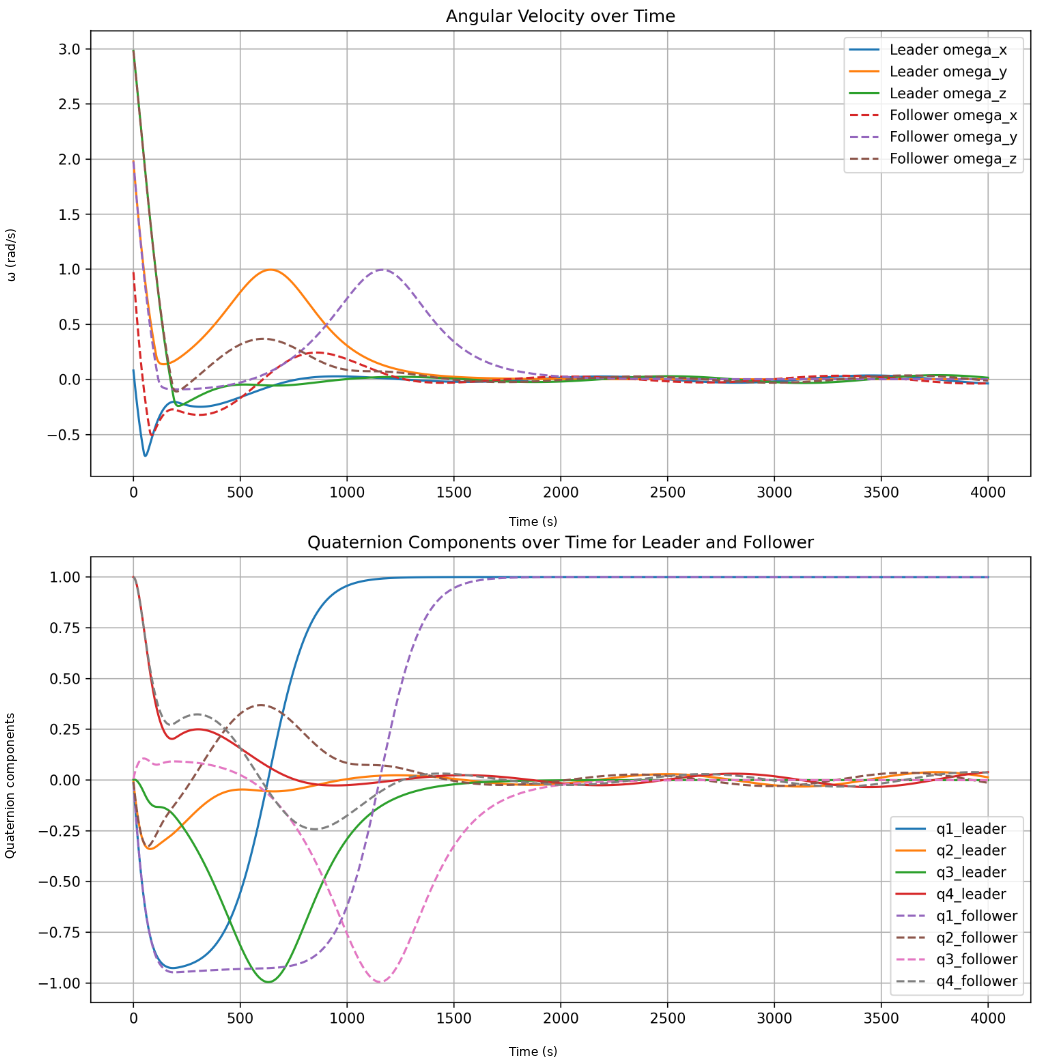}
  \caption{%
  \textbf{AUV rotational dynamics.} 
  The controller drives angular rates to zero and preserves quaternion 
  normalization, achieving stable attitude regulation without chattering 
  despite added-mass and drag uncertainties.
  }
  \label{fig:rot_dynamics}
\end{figure}

\textbf{Results.}
Across runs, the SMC with boundary layers brings both rotational and translational sliding variables to the boundary layer, yielding bounded steady formation error under currents; drag and control inputs remain bounded without high-frequency chatter, and quaternion norm is preserved by RK4 integration. Figure~\ref{fig:trans_dynamics} and figure~\ref{fig:rot_dynamics} show the leader tracks a waypoint and the follower holds a fixed offset under linear–quadratic hydrodynamic drag and Gauss–Markov currents; all states converge smoothly with bounded overshoot.

\section{CONCLUSION}
We presented an explainable AI–enhanced supervisory control framework that couples
(i) a formally verifiable timed–automata supervisor, (ii) robust continuous controllers
(Lyapunov for large–angle convergence and SMC with boundary–layer saturation for
disturbance rejection), and (iii) a learning block that predicts controller gains and
expected performance (energy and error). Using Monte Carlo–driven optimization to
generate training data, the learned predictor enables fast, interpretable gain
selection while the supervisor guarantees safe phase sequencing.

In the space mission (VTXO), the
framework meets tight pointing and alignment requirements with low energy, and the
SMC science controller satisfy the mission constraints. In the AUV
leader–follower case, SMC maintains formation under hydrodynamic
drag and stochastic currents with bounded steady errors. Across both domains the
results show: (1) rapid transient stabilization without chattering, (2) robust
high-precision tracking in the presence of model uncertainty and disturbances, and
(3) interpretable trade-offs between accuracy and resource use.

Looking forward, several extensions are envisioned. First, scaling from
two-agent to many-agent formations will require addressing communication
delays, limited bandwidth, and distributed supervisory logic. Second,
integration with fault detection and recovery modules could allow the
framework to diagnose actuator or sensor degradation and reconfigure
controllers accordingly. Third, tighter coupling with verification and
validation practices may enable contributions to emerging standards for
AI-enhanced autonomy in safety-critical domains (e.g., space missions,
naval robotics, or distributed sensing networks). Finally, heterogeneous
multi-domain autonomy such as aerial–underwater or surface–spacecraft
cooperation—represents a compelling frontier where the framework’s
domain-agnostic structure can be tested. Future efforts target large-scale AUV networks for persistent ocean
monitoring, coordinated inspection of offshore platforms and subsea
pipelines, or nuclear-powered maritime propulsion where autonomy must
remain transparent. 

\bibliographystyle{IEEEtran}
\bibliography{references}

\end{document}